\pdfoutput=1 
\documentclass[10pt,journal,compsoc]{IEEEtran}

\usepackage{hyperref}
\usepackage{url}

\usepackage{booktabs}
\usepackage{graphicx}
\usepackage{lipsum}
\usepackage{microtype}
\usepackage{verbatim}

\usepackage{amsmath}
\usepackage{amssymb}
\usepackage{bm}
\usepackage{todonotes}
\usepackage[capitalize]{cleveref}
\usepackage{multirow}
\usepackage{floatrow}
\usepackage{subfigure}
\usepackage[linesnumbered,ruled,vlined]{algorithm2e}

\ifCLASSOPTIONcompsoc
  \usepackage[nocompress]{cite}
\else
  \usepackage{cite}
\fi

\newcommand{\bx}{\bm{x}}
\newcommand{\by}{\bm{y}}
\newcommand{\bz}{\bm{z}}

\newcommand{\wh}{\textcolor[rgb]{0,0,0}}

\newcommand{\mT}{\mathcal{T}}
\newcommand{\mU}{\mathcal{U}}
\newcommand{\mH}{\mathcal{H}}
\DeclareMathOperator*{\argmax}{arg\,max}

\makeatletter
\DeclareRobustCommand\onedot{\futurelet\@let@token\@onedot}
\def\@onedot{\ifx\@let@token.\else.\null\fi\xspace}
\def\eg{\emph{e.g}\onedot} 
\def\ie{\emph{i.e}\onedot}

\makeatother

\SetKwComment{Comment}{ $\triangleright$ \ }{}
\makeatletter
\renewcommand{\paragraph}{%
	\@startsection{paragraph}{4}%
	{\z@}{0.1em}{-1em}%
	{\normalfont\normalsize\bfseries}%
}
\makeatother

\title{Learning to Impute: A General Framework for Semi-supervised Learning}

\begin{document}

\title{Learning to Impute: A General Framework for Semi-supervised Learning}

\author{Wei-Hong~Li,
        Chuan-Sheng~Foo,
        and~Hakan~Bilen
\IEEEcompsocitemizethanks{\IEEEcompsocthanksitem W.H. Li and H. Bilen are with the VICO group, School of Informatics, University of Edinburgh, United Kingdom.\protect\\
E-mail: w.h.li@ed.ac.uk
\IEEEcompsocthanksitem C.S. Foo is with the Institute for Infocomm Research, A* STAR, Singapore.}
\thanks{Manuscript received April 19, 2005; revised August 26, 2015.}}

\markboth{Journal of \LaTeX\ Class Files,~Vol.~14, No.~8, August~2015}%
{Shell \MakeLowercase{\textit{et al.}}: Bare Demo of IEEEtran.cls for Computer Society Journals}


\maketitle

\begin{abstract}
Recent semi-supervised learning methods have shown to achieve comparable results to their supervised counterparts while using only a small portion of labels in image classification tasks thanks to their regularization strategies.
In this paper, we take a more direct approach for semi-supervised learning and propose learning to impute the labels of unlabeled samples such that a network achieves better generalization when it is trained on these labels.
We pose the problem in a learning-to-learn formulation which can easily be incorporated to the state-of-the-art semi-supervised techniques and boost their performance especially when the labels are limited.
We demonstrate that our method is applicable to both regression and classification problems including facial landmark detection, text and image classification tasks.
\end{abstract}

\section{Introduction}\label{s:intro}

Semi-supervised learning (SSL)~\cite{chapelle2009semi} is one of the approaches to learn not only from labeled samples but also unlabeled ones.
Under certain assumptions such as presence of smooth prediction functions that map data to labels, of low-dimensional manifolds that the high-dimensional data lies~\cite{chapelle2009semi}, SSL methods provide a way to leverage the information at unlabeled data and lessen the dependency on labels.
Recent work~\cite{tarvainen2017mean,miyato2018virtual,berthelot2019mixmatch,xie2019unsupervised} has shown that SSL by using only a small portion of labels can achieve competitive results with the supervised counterparts in image classification tasks (\ie CIFAR10, SVHN).
They built on a variation of the well-known iterative bootstrapping method~\cite{yarowsky1995unsupervised} where in each iteration a classifier is trained on the current set of labeled data, the learned classifier is used to generate label for unlabeled data.
However, the generalization performance of this approach is known to suffer from fitting the model on wrongly labeled samples and overfitting into self-generated labels \cite{tarvainen2017mean}. 
Thus, they mitigate these issues by various regularization strategies. 

While there exist several regularization~\cite{srivastava2014dropout} and augmentation~\cite{zhang2018mixup,devries2017improved} techniques in image recognition problems which are known to increase the generalization performance of deep networks, specific regularization strategies for semi-supervised classification are used to estimate correct labels for the unlabeled data by either encouraging the models to produce confident outputs \cite{lee2013pseudo,grandvalet2005semi} and/or consistent output distributions when their inputs are perturbed \cite{tarvainen2017mean,miyato2018virtual,berthelot2019mixmatch}.
The assumption here is that if a good regularization strategy exists, it can enable the network to recover the correct labels for the unlabeled data and then the method can obtain similar performance with the supervised counterpart when trained on them.
Though this ad-hoc paradigm is shown to be effective, it raises a natural question for a more \emph{direct} approach: Can we instead encourage the network to label the unlabeled data such that the network achieves better generalization performance when trained with them? 

In this paper, we propose a new learning-to-learn method for semi-supervised learning, \textbf{L}earning to \textbf{Impute} (L2I), that can be cast as a bi-level optimization problem to address this question.
Our method involves learning an update rule to label unlabeled training samples such that training our model using these predicted labels improves its performance not only on itself but also on a hold-out set.
Crucially, our method is highly generic and can easily be incorporated to the state-of-the-art methods~\cite{lee2013pseudo,berthelot2019mixmatch} and boost their performance, in particular, in the regime where the number of available labels is limited.
Moreover, our method is not limited to classification problems in computer vision, we show that it can be extended to semi-supervised problems in text classification and also regression tasks where the output space is continuous and achieves significant performance gains.

\section{Related Work}\label{s:relwork}
\subsection{Semi-supervised classification.}
There is a rich body of literature in SSL~\cite{chapelle2009semi} for classification.
Most of the recent work~\cite{lee2013pseudo,tarvainen2017mean,miyato2018virtual,berthelot2019mixmatch,xie2019unsupervised,sohn2020fixmatch,Berthelot2019ReMixMatchSL,Athiwaratkun2019ThereAM} builds on the idea of the bootstrapping technique of ~\cite{yarowsky1995unsupervised} that involves iterative optimization of classifier on a set of labeled data and refinement of its labels for the unlabeled data.
This paradigm is known to overfit on the noisy self-generated labels and thus suffer from low generalization performance.
To alleviate the sensitivity to inaccurate labeling, researchers introduce various regularization strategies.
\cite{grandvalet2005semi} proposes a minimum entropy regularizer that encourages each unlabeled sample to be assigned to only one of the classes with high probability.
\cite{lee2013pseudo} instead follows a more direct approach, use the predicted label with the maximum probability for each unlabeled sample as true-label which is called ``pseudo-label''.

An orthogonal regularization strategy is to encourage a classifier to be invariant to various stochastic factors and produce consistent predictions for unlabeled data when the noise is added to intermediate representations~\cite{srivastava2014dropout} and to input in an adversarial manner~\cite{miyato2018virtual}.
In the same vein, \cite{laine2016temporal,tarvainen2017mean} introduce the $\Pi$-model and Mean-Teacher that are regularized to be consistent over previous training iterations by using a temporal ensembling and teacher/student networks respectively.
Recently \cite{berthelot2019mixmatch} introduced MixMatch algorithm that further extends the previous work by unifying the idea of the consistency regularization and augmentation strategies \cite{zhang2018mixup}. Athiwaratkun et al. \cite{Athiwaratkun2019ThereAM} combine the $\Pi$-model and Mean-Teacher with the Stochastic Weight Averaging \cite{Izmailov2018AveragingWL} while \cite{Berthelot2019ReMixMatchSL,sohn2020fixmatch} extend the MixMatch using existing strong augmentation techniques (\eg RandAugment~\cite{cubuk2019randaugment}, AutoAugment~\cite{cubuk2019autoaugment}, Cutout~\cite{devries2017improved} etc.) to better regularize the network with unlabeled data. As not all unlabeled samples are equal \cite{ren2020not,wang2020meta} develop a learning-to-learn method for learning to weight each unlabeled sample. 
In language domain, \cite{xie2019unsupervised} exploit the idea of consistency regularization and recent data augmentation \cite{wang2018switchout,yu2018qanet} to formulate a semi-supervised learning method for text classification. \cite{miyato2016adversarial} extend the \cite{miyato2018virtual} by applying perturbations to the word embeddings in a recurrent neural network rather than to the original input itself for text classification.
While the recent techniques are shown to be effective in several classification benchmarks, the idea of consistency regularization is still implicit in terms of generalization performance.
Here we take an orthogonal and more direct approach to this approach and learn to impute the labels of the unlabeled samples that improve the generalization performance of a classifier. We also show that our method can be used with the recent SSL techniques. 

\subsection{Semi-supervised regression.}
Some of the recent techniques in semi-supervised classification are shown to be applicable to regression problems. \cite{jean2018semi} have adapted various existing SSL methods such as label propagation, VAT~\cite{miyato2018virtual} and Mean Teacher~\cite{tarvainen2017mean} and studied their performance in regression. The same authors also proposed a Bayesian SSL approach for SSL regression problems which is based on the recent deep kernel learning method~\cite{wilson2016deep} based approach for semi-supervised learning that aims at minimizing predictive variance of unlabeled data. As in the classification task, these methods are typically ad-hoc and does not aim to generate labels for the unlabeled data that are optimized to improve the generalization in regression.


\subsection{Meta-learning.}
Our method is also related to meta-learning~\cite{schmidhuber1987evolutionary,bengio1992optimization,Hospedales2020MetaLearningIN} and inspired from the recent work~\cite{andrychowicz2016learning,finn2017model} where the goal is typically to quickly learn a new task from a small amount of new data. \cite{andrychowicz2016learning,finn2017model} propose a learn gradient through gradient approach to train a model which has a strong generalization ability to unseen test data and can be easily adapted to new/unseen tasks. \cite{sung2018learning} introduce the relation network to learn an additional metric such that the learned model's feature embedding can be generalized for unseen data and unseen tasks. 
\cite{ren2018learning} adopt the meta learning to learn the weight of samples to tackle the sample imbalance problem.
\cite{lorraine2018stochastic,mackay2019self} employ meta learning for automatically learning the hyper-parameters of the deep neural networks. 
Meta-learning has recently been applied to unsupervised learning \cite{hsu2018unsupervised} and SSL for few shot learning \cite{ren2018meta}. 
\cite{ren2018meta} adapts the prototypical network to use unlabeled examples when producing prototypes, enabling the prototypical network to exploit those unlabeled data to assist few shot classification. 
\cite{Antoniou2019critique} propose to learn a label-free loss function, parameterized as a neural network that enables the classifier to leverage the information from a validation set and achieves better performance in few-shot learning.
\cite{sun2019learning} propose a meta-learning technique to initialize a self-training model and to filter out noisy pseudo labels for semi-supervised few-shot learning.
Similarly, our work also builds on the ideas of optimizing for improving the generalization for unseen samples.
However, in contrast to the existing meta-learning methods that is proposed for few-shot learning problems, we focus on semi-supervised learning in general classification and regression problems where the number of samples are not limited to few.


\begin{figure*}[t]
	\centering
	\includegraphics[width=.9\linewidth]{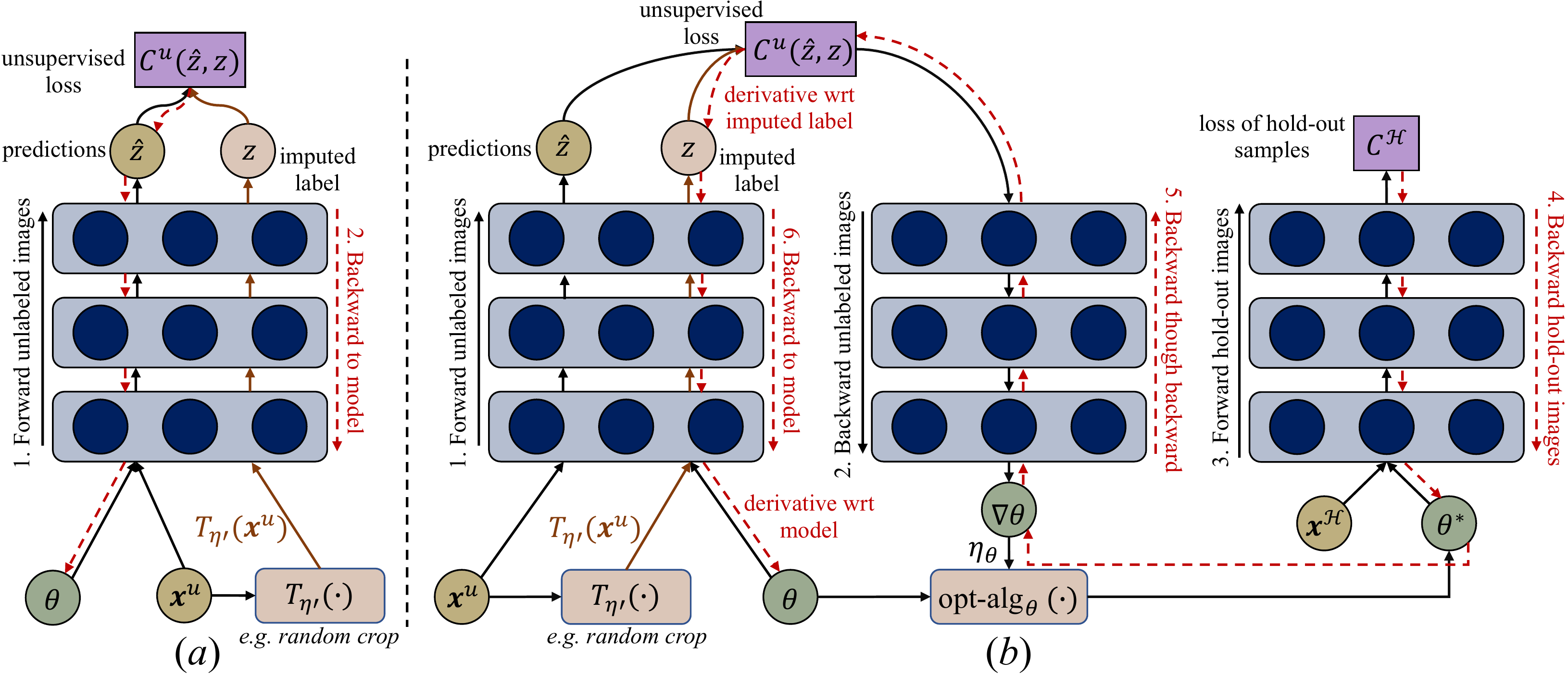}  
	\vspace{-.3cm}
	\caption{Computation graph of (a) consistency based semi-supervised learning and (b) our algorithm on imputing unlabeled samples in a deep neural network. Note that the illustration shows three instances of the same network and our method uses a single neural network. Best seen in color.}
	\label{fig:diagram}
\end{figure*}

\section{Preliminaries for SSL with consistency regularization}\label{sec:sslreview}
Consider a dataset $\mathcal{D}$ of $N$ samples $\{(\bx_1,\by_1),(\bx_2,\by_2),\dots,(\bx_{N},\by_{N})\}$ where each sample $\bx$ is labelled with $\by$.
The dataset is further split into a training $\mT$ and hold-out set $\mH$ with $|\mT|$ and $|\mH|$ samples respectively.
We also have a set of unlabeled samples $\mU=\{\bx^{u}\}_{i=1,\dots, |\mU|}$.
Our goal is to learn $\Phi_\theta$, a model (\eg a deep neural network) with parameters $\theta$ to correctly predict label of an unseen sample $\bx$ as $\by = \Phi_\theta(\bx)$.
In the supervised case, we aim to minimize the expected loss on the training set:
\begin{equation}\label{eq:costT}
C^{\mT}=\sum_{\bx,\by\in \mT} \ell(\Phi_\theta(\bx),\by)
\end{equation} where $\ell$ is a task specific loss, such as a soft-max followed by a cross-entropy loss for classification or squared loss for regression tasks.

We are also interested in exploiting the information in the unlabeled samples.
As their ground-truth labels are not available, many recent semi-supervised methods~\cite{lee2013pseudo,tarvainen2017mean,berthelot2019mixmatch,sohn2020fixmatch} employ a regularization term on the unlabeled samples that encourages the model to produce consistent predictions when its inputs are perturbed.
In the context of image classification, commonly applied perturbations include geometric transformations (rotation, cropping, scaling) and color jittering.

Without loss of generality, the regularization term or \emph{consistency loss}~$\ell^u$ can be written as a function that computes the difference (defined by a separate function $d$) between the output of the network $\Phi_\theta$ for a (transformed) unlabeled sample $\bx^u$ and an imputed label $\bz^{(\bx^u)}$
\[
\ell^u(\bx^u, \bz^{(\bx^u)}; \Phi_\theta) = d(\Phi_\theta(T_{\eta}(\bx^u)),\bz^{(\bx^u)}),
\]
with $T_\eta$ being a random transformation (perturbation) on $\bx$ parameterized by $\eta$; common choices of $d$ include cross-entropy and Euclidean distance, as further discussed below. The empirical loss on the unlabeled set is then
\begin{equation}\label{eq:costU}
C^{\mU}(Z)=\sum_{\bx^u\in \mU}
\ell^u(\bx^u, \bz^{(\bx^u)}; \Phi_\theta).
\end{equation} where $Z$ is the set that contains the imputed labels for  $\bx^u\in\mU$. 
The imputed labels can further be generally defined as $\bz^{(\bx^u)} = \psi(\bx^u)$ for an imputing function $\psi$.
As an example, we can let $\psi=\Phi_\theta(T_{\eta'}(\bx^u))$ such that $\bz^{(\bx^u)}$ is obtained by computing the network's output for another random transformation of the sample $\bx^u$.
In this case, minimizing $C^{\mU}$ encourages the network to produce predictions that are invariant to the transformations used.
We briefly review commonly used imputing functions $\psi$ and difference functions $d$ in the following:


\vspace{0.05cm}
\noindent \textbf{Mean-teacher (MT) \cite{tarvainen2017mean}} involves two deep neural networks, denoted as student and teacher, where the weights of the teacher model $\theta'$ are an exponential moving average of the student model's weights $\theta$. 
Formally, at training iteration $t$, $\theta'_{t}=\alpha \theta'_{t-1}+ (1-\alpha)\theta_{t}$ where $\alpha$ is a smoothing coefficient hyperparameter. The labels for unlabeled samples are imputed with the teacher model: $\psi(\bx^u)=\Phi_{\theta'}(T_{\eta'}(\bx^u))$, and 
$d$ is the Euclidean distance between predictions of the two models so the consistency loss $\ell^u = \lvert\lvert \Phi_\theta(T_{\eta}(\bx^u))-\bz^{(\bx^u)}\rvert\rvert^2$.
In this case, the student is trained to not only be robust to various transformations but also to be consistent with the teacher.

\vspace{0.05cm}
\noindent \textbf{MixMatch (MM) \cite{berthelot2019mixmatch}} imputes labels by applying a sharpening function $\psi(\bx^u)=\texttt{sharpen}\left(\frac{1}{K}\sum_{i=1}^{K}\Phi_{\theta}(T_{\eta_i}(\bx^u)),\beta\right)$
where \texttt{sharpen} is a softmax operator that takes in the averaged prediction over $K$ augmentations of $\bx^u$ and $\beta$ is the temperature hyperparameter.
Similar to MT, MM also uses an Euclidean distance for $d$.
Note that MM also uses a recent augmentation strategy, MixUp \cite{zhang2018mixup} that mixes both labeled and unlabeled examples with label guesses as a transformation.
We refer the interested reader to \cite{berthelot2019mixmatch} for more details.

\vspace{0.05cm}
\noindent \textbf{FixMatch (FM) \cite{sohn2020fixmatch}} imputes the labels for a given unlabeled sample $\bx^u$ as follows: it first generates a soft-label by computing the network output for an augmentation of the unlabeled sample, picks the class with the maximum predicted probability, and then generates an one-hot ``pseudo-label'' vector where all the entries are zero for all categories but the maximum one, \ie $\bz^{(\bx^u)}=\argmax\left(\Phi_\theta(T_{\eta'}(\bx^u))\right)$ where $\eta'$ is a random transformation. 
The authors employ the cross-entropy loss function for $d$ on the network output of a ``strong'' augmentation (\eg RandAugment \cite{cubuk2019randaugment}) of the unlabeled sample and its imputed label.
This strategy encourages the learned model to be invariant to the strong jittering.

With the imputed labels, model parameters $\theta$ can then be determined by minimizing a linear combination of the two loss functions weighted by $\lambda$
\begin{equation}
\theta^{\ast}(Z) =\arg\min_{\theta} \; C^{\mT}+\lambda C^{\mU}(Z).
\label{eq:opt1}
\tag{OPT1}
\end{equation}
Next we consider the network parameters as a function of the imputed labels and parameterize $\theta^{\ast}$ by $Z$, as the optimum solutions to \ref{eq:opt1} depend on the set of imputed labels used.

\begin{algorithm*}[t]
	\caption{Pseudo-code of our method L2I.}
	\label{algorithm:option12}
	\DontPrintSemicolon
    \SetAlgoLined
	\footnotesize
	\KwIn{$\mT$, $\mU$, $\mH$ \Comment*[f]{ training set, unlabeled and hold-out set resp.}}
	\textbf{Required}: Initial model function $\Phi_{\theta^0}$ and imputing function $\psi_{\theta^0}$.
	
	\For{$t = 0, \cdots, K-1$}
	{
		 $B^{\mT}\sim\mT$, $B^{\mU}\sim\mU$, $B^{\mH}\sim\mH$  \Comment*[f]{randomly sample minibatches}.\\

	    $ Z = \{\psi_{\theta^t}(T_{\eta'}(\bx^{u}))\;|\;\bx^u\in B^{\mU}\}$ \Comment*[f]{ impute labels for unlabeled samples}\\
	    
	    ${C(\theta^t,Z)}=\sum_{i=1}^{|B^{\mT}|}\ell(\Phi_{\theta^t}(\bx_i), y_i)+\lambda \sum_{i=1}^{|B^{\mU}|}\ell^u(\Phi_{\theta^t}(\bx_i^{u}), \bz^{(\bx^u_i)})$ \\
	    $\hat{\theta}^{t} = \texttt{opt-adam}_{\theta}(C(\theta^t,Z))$\Comment*[f]{update the model by using Adam optimizer}\\	    
%
	    
	    $ Z = \{\psi_{\hat{\theta}^{t}}(T_{\eta'}(\bx^{u}))\}$ where $\bx^u\in B^{\mU}$ \Comment*[f] {re-impute labels using the updated model}\\
	    
	    ${C(\hat{\theta}^t,Z)}=\sum_{i=1}^{|B^{\mT}|}\ell(\Phi_{\hat{\theta}^t}(\bx_i), y_i)+ \sum_{i=1}^{|B^{\mU}|}\ell^u(\Phi_{\hat{\theta}^t}(\bx_i^{u}), \bz^{(\bx^u_i)})$\\
	    
	    $\theta^{\ast} \leftarrow \texttt{opt-SGD}(C(\hat{\theta}^t,Z))$\Comment*[f]{inner-loop optimization}\\

	    ${C^\mH(\theta^{\ast})}=\sum_{i=1}^{|B_t^{\mH}|}\ell(\Phi_{\theta^{\ast}}(\bx_i), y_i)$ \Comment*[f]{evaluate on hold-out set}\\
	    
		   
		${\theta}^{t+1} \leftarrow	\texttt{opt-adam}_{\theta}(C^{\mH}(\theta^{\ast}))$\Comment*[f]{outer-loop optimization: update the model}
	}
	\KwOut{$\theta^{K}$}
\end{algorithm*}

\section{Method}\label{s:method}
State-of-the-art methods focus on engineering imputation strategies to obtain better generalization performance when training models with
\ref{eq:opt1} \cite{laine2016temporal,tarvainen2017mean,berthelot2019mixmatch}.
However, this process is highly manual 
as it relies on trial and error.
We instead propose a more direct approach, \emph{to learn $Z$ from data} such that $\theta^{\ast}(Z)$ generalizes well. 

To this end, we define $Z$ as the output of $\psi$, which in turn is typically a function of $\Phi$ (see \cref{sec:sslreview}), and thus also influenced by $\theta$. 
Hence, our goal is to find the parameters $\theta$ that give us the best performance on a hold-out set through its influence on the imputed labels $Z$:
\begin{equation}\label{eq:opt2}
	\begin{split}
	{} & \min_{\theta} \; \; C^{\mH}(\theta^{\ast}) \\  
	\text{s.t.}\;\;  \theta^{\ast} = \arg & \min_{\theta'} \; C^{\mT}(\theta')+\lambda C^{\mU}(\theta', Z(\theta))
	\end{split}
	\tag{OPT2}
\end{equation} where $\theta$ denotes the parameters that are used to impute the missing labels.
Note that our model does \emph{not} require two sets of parameters, here we use $\theta$ and $\theta'$ to show that the outer loop optimization optimizes $C^{\mH}$ w.r.t. $\theta$ through the imputed labels $Z$.


\noindent \textbf{Solving \ref{eq:opt2}.}
Standard approaches to solving \ref{eq:opt2} employ implicit differentiation, which involves solving the inner optimization for $\theta^*$ and an associated linear system to high precision to recover the gradient for $\theta$ (see \cite{Do2008,samuel2009learning,domke2012generic} for details). 
However, this approach is challenging to implement in practice for deep neural networks, due to the high computational cost of training networks to convergence even for a single gradient; in the event that solutions are not obtained with sufficient precision, the resulting gradients become too noisy to be useful \cite{domke2012generic}.

As such, we adopted the back-optimization approach of \cite{domke2012generic} which instead defines $\theta^*$ as the result of an incomplete optimization. Formally, 
\begin{equation}\label{eq:opt1a}
\theta^{\ast}(Z(\theta)) =\texttt{opt-alg}_{\theta} \; (C^{\mT}+ \lambda C^{\mU}(Z))
\end{equation} where \texttt{opt-alg} denotes a specific optimization procedure with a fixed number of steps, which sets a trade-off between accuracy and computational load.
We wish to solve \ref{eq:opt2}, even in cases where the optimization in \cref{eq:opt1a} has not converged.
As we experiment with deep neural networks, we use multiple steps of SGD for \texttt{opt-alg} for simplicity; it is also possible to employ other optimizers in this framework (like SGD with momentum). We can then compute the gradient over $\theta$ through $Z$ using automatic differentiation by explicitly unrolling the optimization steps over $\theta^*$.

\vspace{0.05cm}
\noindent \textbf{Algorithm.} 
\Cref{fig:diagram} depicts an overview of previous SSL methods in (a) and our method in (b).
Recent SSL methods learn to fit their models to the imputed labels under various input transformations, however, invariance to input transformations alone does not guarantee that imputed labels are correct and results in better generalization.
In contrast, our method performs a look-ahead by simulating a training step on the train and unlabeled set with the imputed labels and determine the second order effects of updating the model through the imputed labels.
This strategy helps to reduce the confirmation bias that happens through fitting the model to the imputed labels, and also enables our method to impute labels that improves the performance in a hold-out set.
In particular, recent SSL methods that optimize a consistency loss ($C^{\mU}$) to update its parameters $\theta$ (\ie step 1, 2 in \Cref{fig:diagram}(a)) for each unlabelled image $\bx^u$ and its imputed label $\bz^{(\bx^u)}$, whereas our method utilizes the update to produce an intermediary model $\theta^\ast$ which is evaluated on a hold-out set (\ie step 1, 2, 3 in \Cref{fig:diagram}(b)), then computes an update for $\theta$ by unrolling the gradient graph over $\theta^\ast$ (\ie step 4, 5, 6 in \Cref{fig:diagram}(b)).

We also give a more detailed description of the optimization procedure in Alg.~\ref{algorithm:option12}.
Given $\mT$, $\mU$ and $\mH$, we first randomly sample a mini-batch from each set, denote them as $B^{\mT}$, $B^{\mU}$ and $B^{\mH}$ respectively at each iteration $t$.
We then initialize $\bz$ for the unlabeled batch $B^{\mU}$ by using $\psi(\cdot)$.
Note that we evaluate our method for various $\psi$ selection in \cref{s:exp}.
Next we optimize the model parameters with the loss $C^{\mT}+\lambda C^{\mU}$ by using the Adam optimizer~\cite{kingma2014adam} and obtain $\hat{\theta}^t$. 
We re-impute $\bz$ with $\hat{\theta}^t$ and run the inner-loop by minimizing $C^{\mT} + \lambda C^{\mU}$, obtaining $\theta^{\ast}$ for outer-loop optimization.

For the inner-loop, we use a SGD optimizer with fixed number of steps for \texttt{opt-alg} in \cref{eq:opt1a} to obtain $\theta^{\ast}$ by using \cref{eq:opt1a}.
For one step SGD, \texttt{opt-alg} is simply given by $\theta^{\ast} = \hat{\theta}^t - \eta_{\theta} \nabla_\theta (C^{\mT}+ C^{\mU})$ where $\nabla_\theta$ is the gradient operator and $\eta_{\theta}$ is the learning rate.

For the outer-loop optimization, we wish to find $Z$ via $\theta$ that minimizes $C^{\mH}(\theta^{\ast})$ in \ref{eq:opt2}.
Now we apply the chain rule to the objective function of \ref{eq:opt2} to obtain
\begin{equation}\label{eq:updopt}
	\frac{\partial{C}^\mH}{\partial \hat{\theta}^{t}} = 
	\frac{\partial C^{\mH}}{\partial \Phi_{\theta^{\ast}}} \frac{\partial \Phi_{\theta^{\ast}}}{\partial \theta^{\ast}} \frac{\partial \theta^{\ast}}{\partial \bz }
	\frac{\partial \bz}{\partial \hat{\theta}^{t}}
\end{equation} where ${\partial \theta^{\ast}}/{\partial \bz}={\partial \hat{\theta}^t}/{\partial \bz}- \eta_{\theta} \nabla_{\bz}\nabla_\theta C^{\mU}$ when \texttt{opt-alg} is an one-step SGD optimizer.
We then use the gradient (${\partial{C}^\mH}/{\partial \theta} $) for \ref{eq:updopt} to update the model parameters. 

\vspace{0.1cm}
\noindent \textbf{Discussion.} So far, we have modelled the imputed labels $Z$ as the output of $\psi$, thus a function of $\Phi$.
Alternatively, one can treat $\bz$ as learnable parameters and minimize the hold-out loss in \cref{eq:opt2} by optimizing them.
Compared to the proposed method, this strategy would require two steps, optimizing the labels and then training the model on the unlabeled set with these learned labels and thus would be more implicit.
We also analyse this strategy in the next section.

\vspace{0.05cm}
\noindent \textbf{Analysis.} We attempt to provide insight into the behavior of our algorithm by analytically computing the update rule \ref{eq:updopt} for an one-layer network in the semi-supervised binary classification setting. 
Here, $\Phi_{\theta}(\bx)=\sigma(\theta^\top \bx)$ where $\sigma$ is the sigmoid function.
In this case, for a given unlabeled sample $\bx^u$, its associated imputed label $z$ and a set of hold-out samples, the update rule \ref{eq:updopt} for $\theta$ is: 
\begin{equation}
\begin{aligned}
\frac{\partial C^{\mH}}{\partial \bz}\frac{\partial \bz}{\partial \theta} &= \eta_{\theta}\sum_{\bx,y\in\mH}((\sigma({\theta^*}^\top\bx)-y)\bx^\top\bx^u) \frac{\partial \bz}{\partial \theta}
\end{aligned}
\label{eq:sod}
\end{equation} where $\theta^{\ast}$ is obtained with \ref{eq:opt1}, $y\in\{0,1\}$, $\eta_\theta$ is the learning rate for $\theta$.
The update rule is proportional to the similarity (dot product) between the unlabeled and hold-out samples, and to the loss between the prediction and the ground-truth labels. 
Intuitively, the algorithm updates the model weights to impute the labels towards the majority class of the incorrectly predicted samples on the hold-out set, weighted by the similarity of the unlabeled sample to the respective hold-out sample; correctly predicted hold-out samples do not contribute to the update. 
We provide the details of the derivation for binary classification and also regression in the supplementary.

\vspace{0.05cm}
\noindent \textbf{Learning to impute efficiently.}
Training our method requires an extra back-propagation step (see line 6 in Alg.~\ref{algorithm:option12}) over the standard semi-supervised learning training as well as a second-order derivative computation on unlabeled samples, which slows down our training significantly.
Computing the second order derivative involves a Hessian matrix computation which is computationally expensive in case of large networks and datasets.

Inspired by the above analysis for the one-layer network, we consider our model $\Phi$ as a multi-layered feature extractor network followed by a linear classifier.
To speed up our training, we compute the derivative only for parameters in the last layer and use the derived update rule as in~\cref{eq:sod} that approximates \cref{eq:updopt} using the similarity between the features of the unlabeled and hold-out samples.
$\frac{\partial C^{\mH}}{\partial \partial \theta}$ is approximated as follows:
\begin{equation}
\begin{aligned}
\eta_{\theta}\sum_{\bx,y\in\mathcal{H}}(\Phi_{\theta^{\ast}}(\bx)-y)\texttt{vec}(\Phi_{\theta^{\ast}}^{f}(\bx))^\top\texttt{vec}(\Phi_{\theta}^{f}(\bx^u))\frac{\partial \bz}{\partial \theta}
\end{aligned}
\label{eq:approx}
\end{equation} where the parameters of a deep neural network~$\theta$ is decomposed into a feature encoder and classifier that are parameterized by $\theta_f$ and $\theta_c$ respectively. 
$\Phi_{\theta^{\ast}}(\bx)$ is the prediction (posterior) of a hold-out sample $\bx$ by $\theta^{\ast}$, $\Phi_{\theta^{\ast}}^{f}$ is the feature encoder and $\texttt{vec}(\cdot)$ is the vectorization operator. 
We use the approximated version of our method in all the experiments and also show that the approximation significantly speeds up at train time without any significant drop in performance \cref{s:exp}. 



\section{Experiments}\label{s:exp}

\subsection{Experiment setup}
We evaluate our method on multiple benchmarks for regression and classification problems, and also analyse multiple hyperparameters and design decisions.
In all experiments, we use the validation sets \emph{only} for tuning hyperparameters such as inner, outer-loop learning rates.
The hold-out set ($C^\mH$) is sampled from the training set in all our experiments as in \cite{rajeswaran2019meta,finn2017model,vinyals2016matching,sung2018learning}.
As depicted in Alg.~\ref{algorithm:option12}, the algorithm utilizes three minibatches -- $B^\mT$, $B^\mU$, $B^\mH$ -- at each training iteration.
While the hold-out minibatch $B^\mathcal{H}$ can be obtained from a separate validation set, we instead randomly sample two minibatches for $B^\mathcal{T}$ and $B^\mathcal{H}$ from the \emph{same training set} at each iteration, to avoid using more labeled data than the baselines. 
This strategy does not suffer from overfitting to the hold-out set, as i) the hold-out loss in \ref{eq:opt2} is optimized for a single randomly sampled minibatch at each iteration, ii) we apply data augmentation to both unlabeled and hold-out batches.
We study the effect of the hold-out batch size and separate hold-out set in the supplementary and observe that the performance our method is not sensitive to the hold-out batch size and using a separate hold-out set can further improve the performance.
\footnote{Our implementation in PyTorch is available at \url{https://github.com/VICO-UoE/L2I/}.}

\subsection{Semi-supervised regression}

\vspace{0.05cm}
\noindent \textbf{AFLW.}
We first evaluate our method on a regression task -- predicting the location of 5 facial landmarks from face images -- using the Annotated Facial Landmarks in the Wild (AFLW) dataset~\cite{koestinger2011annotated,zhang2015learning}.
We use the official train (10,122 images) and test (2991 images) splits, randomly pick 10\% (1012 images) of samples of the original training set as the validation set \emph{only} for hyperparameter tuning and early stopping, and use the rest of this data as labeled and unlabeled data in our experiments.
We evaluate the baselines and our method for various portions of labeled training images and report the standard Mean Square Error (MSE) normalized by the inter-ocular distance as in \cite{zhang2015learning}.

\begin{table}
	\resizebox{1.0\textwidth}{!}
	{
		\begin{tabular}{lccc}
			\toprule
			
			
			\#labels  & 91 (1\%) & 182 (2\%) & 455 (5\%) \\
			\midrule
			SL & 16.32 $\pm$ 0.14 & 14.21 $\pm$ 0.02 & 11.81 $\pm$ 0.00  \\
			\midrule
			PL~\cite{lee2013pseudo} & 12.40 $\pm$ 0.03 & 10.77 $\pm$ 0.05 & 9.16 $\pm$ 0.04 \\
			{\bf PL-L2I (Ours)} & {\bf 11.72 $\pm$ 0.04} & {\bf 10.12 $\pm$ 0.06} & {\bf 8.94 $\pm$ 0.03} \\ 
			\midrule
			MT~\cite{tarvainen2017mean} & 12.80 $\pm$ 0.00 & 11.43 $\pm$ 0.02 & 9.46 $\pm$ 0.01 \\
			{\bf MT-L2I (Ours)} & {\bf 11.53 $\pm$ 0.02} & {\bf 10.32 $\pm$ 0.08} & {\bf 8.89 $\pm$ 0.08} \\
			\bottomrule
		\end{tabular}%
	}
	\caption{Testing mean square error on AFLW for the facial landmark detection task. Error rate of the supervised learning method trained on the whole training is 7.71 $\pm$ 0.04.}
	\label{tab:AFLW}%
\end{table}%




\vspace{0.05cm}
\noindent \textbf{Experiment 1.}
For the regression task, we adopt the TCDCN backbone architecture in \cite{zhang2015learning} and SGD optimizer, use standard data augmentation. 
We train all methods for $22500$ steps and set the initial learning rate to $0.03$ and reduce it by 0.1 at 15,000 steps. 
The momentum and weight decay are set to 0.9 and 0.0005 resp.  
We use supervised learning (SL), Pseudo Labeling (PL)~\cite{lee2013pseudo} and MeanTeacher (MT)~\cite{tarvainen2017mean} as the baselines and also apply our method to both PL and MT.
Note that MixMatch (MM)~\cite{berthelot2019mixmatch} is not applicable to this task, as mixing up two face images doubles the number of landmarks.
\Cref{tab:AFLW} depicts the results for the baselines and ours in mean error rate.

\begin{table*}
\centering
		\resizebox{0.9\textwidth}{!}
		{
		\begin{tabular}{lccc|ccc}

		& \multicolumn{3}{c}{CIFAR10} & \multicolumn{3}{c}{CIFAR100} \\
		\toprule
		
		\#labels           & 250   & 500    & 1000   & 500 & 1000 & 2000 \\
		\midrule
		$\Pi$ model~\cite{laine2016temporal} & 53.02 $\pm$ 2.05 & 41.82 $\pm$ 1.52 & 31.53 $\pm$ 0.98  & -\ - & -\ - & -\ - \\
		PL~\cite{lee2013pseudo} & 49.98 $\pm$ 1.17 & 40.55 $\pm$ 1.70 & 30.91 $\pm$ 1.73  & -\ - & -\ - & -\ - \\
		MixUp~\cite{zhang2018mixup} & 47.43 $\pm$ 0.92 & 36.17 $\pm$ 1.36 & 25.72 $\pm$ 0.66  & -\ - & -\ - & -\ - \\
		VAT~\cite{miyato2018virtual} & 36.03 $\pm$ 2.82 & 26.11 $\pm$ 1.52 & 18.68 $\pm$ 0.40  & -\ - & -\ - & -\ - \\
		\midrule
		MT~\cite{tarvainen2017mean}           & 37.75 $\pm$ 0.12 & 27.89 $\pm$ 0.52  & 19.15 $\pm$ 0.15  & 74.42 $\pm$ 0.31 & 61.94 $\pm$ 1.11 & 52.01 $\pm$ 0.24  \\
		{\bf MT-L2I (Ours)} & {\bf 25.97 $\pm$ 1.71} & {\bf 21.13 $\pm$ 0.32} & {\bf 14.76 $\pm$ 0.50}  & {\bf 70.58 $\pm$ 0.19} & {\bf 58.96 $\pm$ 0.14} & {\bf 51.95 $\pm$ 0.06}  \\
		\midrule
		MM~\cite{berthelot2019mixmatch}           & 11.71 $\pm$ 0.44 & 10.41 $\pm$ 0.11   & 9.74 $\pm$ 0.30 & 74.67 $\pm$ 0.18 & 60.75 $\pm$ 0.16 & {\bf 50.26 $\pm$ 0.06} \\
		{\bf MM-L2I (Ours)} & {\bf 10.94 $\pm$ 0.16} & {\bf 10.36 $\pm$ 0.05}  & {\bf 9.48 $\pm$ 0.10}   & {\bf 68.37 $\pm$ 0.24} & {\bf 57.83 $\pm$ 0.39} & 50.33 $\pm$ 0.01 \\
		\midrule 
		FM~\cite{sohn2020fixmatch}      & 5.70 $\pm$ 0.04 & 5.78 $\pm$ 0.04 & 5.55 $\pm$ 0.07 & 48.32 $\pm$ 0.23 & 40.68 $\pm$ 0.10 & 35.52 $\pm$ 0.06 \\
		{\bf FM-L2I (Ours)} & {\bf 5.56 $\pm$ 0.03} & {\bf 5.70 $\pm$ 0.02} & {\bf 5.38 $\pm$ 0.02} & {\bf 44.97 $\pm$ 0.13} & {\bf 39.21 $\pm$ 0.10} & {\bf 35.45 $\pm$ 0.10} \\
		\bottomrule
	\end{tabular}%
	}
		\vspace{-.3cm}
		\caption{Test error (\%) in CIFAR-10 \& -100 using WideResNet-28-2.}
		\label{tab:cifar-soa}%
	\end{table*}

\begin{figure}
\centering
\includegraphics[width=0.95\linewidth]{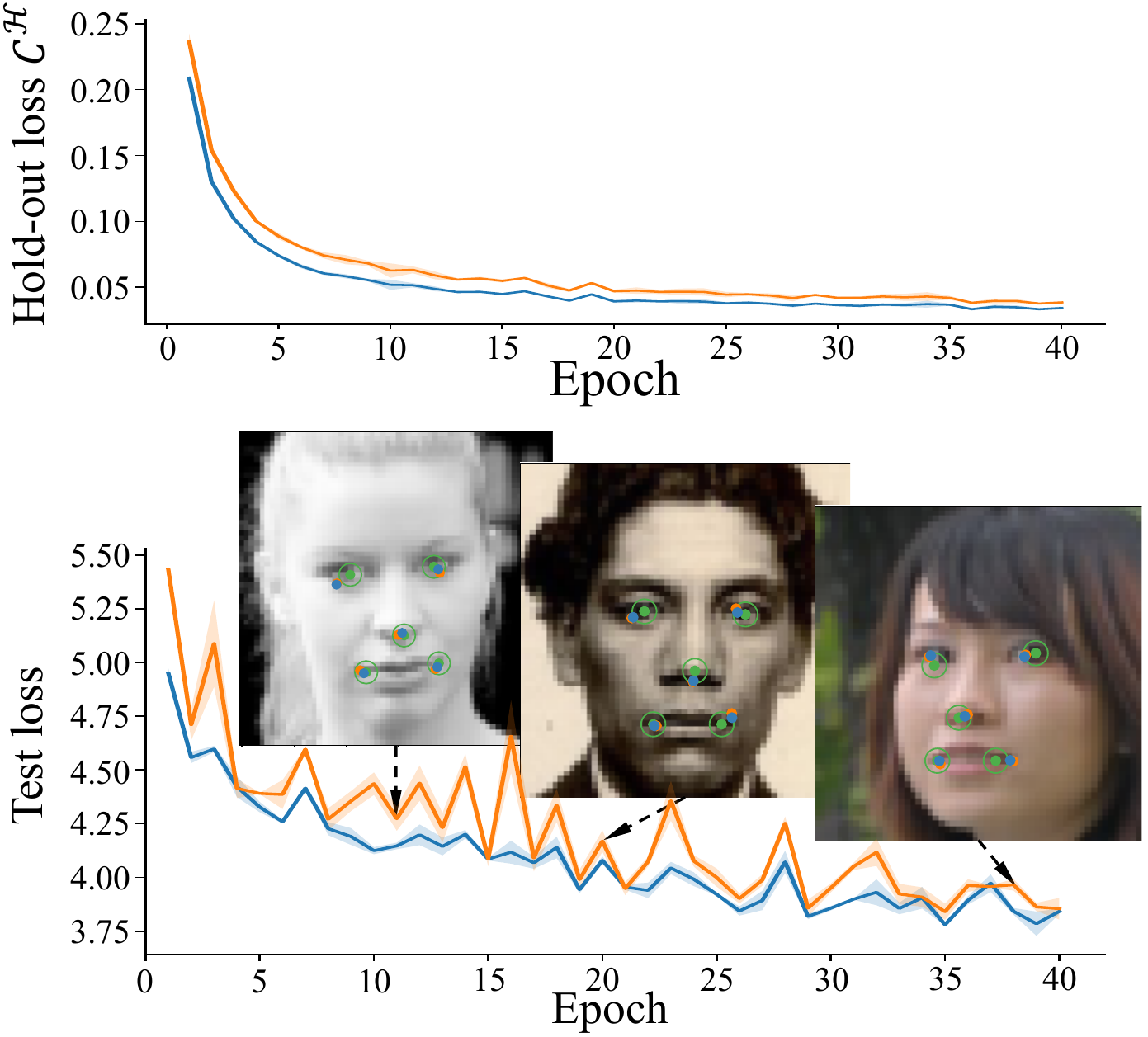}
\caption{Effect of update on labels on facial landmarks detection. {\color{cyan} Cyan curves} are the loss of the model after the update while {\color{orange} orange curves} are the model's loss before the update. Best seen in color.}
\label{fig:alfwcurves}
\end{figure}

\begin{figure}
\centering
\includegraphics[width=0.95\linewidth]{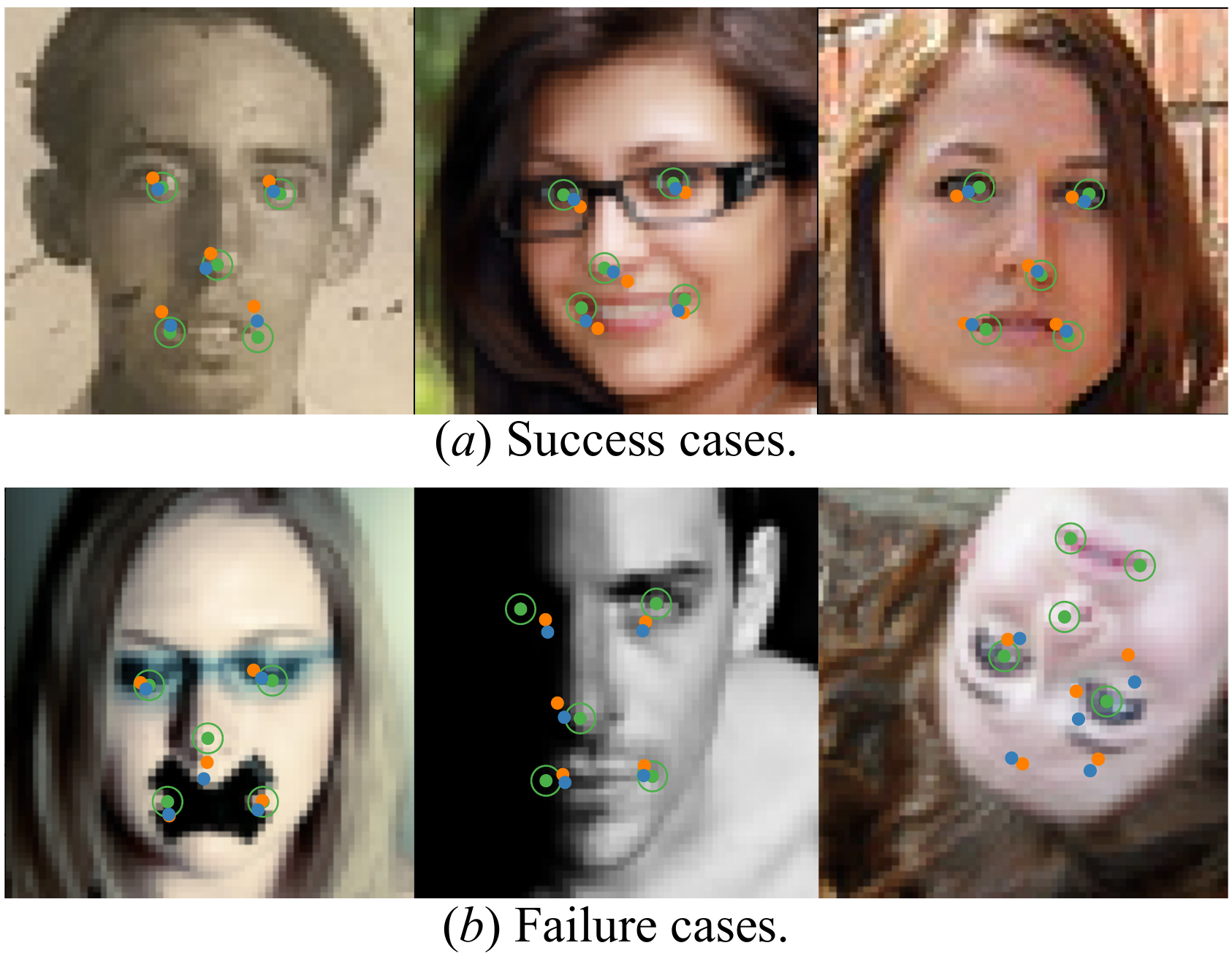}
\caption{Success (a) / failure cases (b) on AFLW for facial landmark detection. {\color{green} Green points with circles}, {\color{cyan} cyan points} and {\color{orange} orange points} are ground-truth landmarks, predictions of our method and Mean Teacher respectively. Best seen in color.}
\label{fig:visualalfw} 
\end{figure}

We observe that the supervised learning on 1\% of the labels is very challenging and obtains only 16.32\% which is on par with the performance of simply taking the mean of each facial landmark over all training samples (16.58\%).
Leveraging the unlabeled face images improves PL as well as MT significantly improves over SL.
Our method achieves consistent improvement over PL and MT for different portions of labels by refining the estimated landmarks of PL and MT on the unlabeled images.

We also analyze the effect of updates on the landmarks in training in \cref{fig:alfwcurves}; the top plot depicts the regression loss on the hold-out batch before and after the update and shows that updating the imputed landmarks in the unlabeled images leads to a better accuracy on the hold-out samples; the bottom plot depicts the test loss for the same models and shows that the updated model (blue curve) does not overfit to the hold-out set and generalizes better to the test images.
We also visualize the updates on the landmarks on the example test images and observe that the updates help them to get closer to the ground-truth ones.

We also illustrate success/failure cases of our method on the test images in \cref{fig:visualalfw} and depict the ground-truth, predicted landmarks by MT and our method when trained with 1\% of the labeled data.
The difference is visually significant and our method outputs more accurate landmarks than MT in the successful ones.
The bottom row shows the cases of extreme pose variation and occlusion where both MT and ours fail to achieve accurate predictions.

\subsection{Semi-supervised classification}

\vspace{0.05cm}
\noindent \textbf{Image Classification on CIFAR-10 \& -100.}
We also evaluate our method on the image classification benchmarks~\cite{krizhevsky2009learning} that are common benchmarks for both supervised and semi-supervised classification.
Both datasets contain 50,000 and 10,000 training and testing samples respectively.
In our experiments, we strictly follow the training and testing protocol for semi-supervised learning in~\cite{oliver2018realistic,berthelot2019mixmatch} -- 5000 of training samples are used as validation data, the remaining 45,000 training samples are split into labeled and unlabeled training sets.
As in \cite{oliver2018realistic}, we randomly pick $|\mathcal{T}|$ samples as labeled data and the rest ($|\mathcal{U}|=45,000-|\mathcal{T}|$ samples) are unlabeled data.
We report results for multiple training data-regimes $|\mathcal{T}|=250, 500, 1000$ on CIFAR-10 and $|\mathcal{T}|=500, 1000, 2000$ on CIFAR-100.
Note that we do \emph{not} use the validation data in training of the model parameters but only for hyperparameter selection.


\vspace{0.05cm}
\noindent \textbf{Experiment 2.} Here, we show that our method performs well on image classification tasks and it can be incorporated to the state-of-the-art methods such as MT~\cite{tarvainen2017mean}, MM~\cite{berthelot2019mixmatch}, and FM~\cite{sohn2020fixmatch} that use a sophisticated backbone networks, augmentation and regularization strategies, and also boost their performance.
For this experiment, we follow the implementation of \cite{berthelot2019mixmatch,sohn2020fixmatch} -- we adopt a competitive backbone, WideResNet-28-2, use the Adam optimizer along with standard data augmentation techniques (see supplementary for more details). 

\Cref{tab:cifar-soa} depicts the classification error rate for the several state-of-the-art techniques including $\Pi$ model~\cite{laine2016temporal}, PL~\cite{lee2013pseudo}, MixUp~\cite{zhang2018mixup}, VAT~\cite{miyato2018virtual}, MT~\cite{tarvainen2017mean}, MM~\cite{berthelot2019mixmatch} and FM~\cite{sohn2020fixmatch}.
Note that some methods do not report results on CIFAR-100; all the results except those for MT, MM and FM are taken from the original papers. 
As we built our methods on top of MM, MT, and FM to verify that our method can be incorporated to state-of-the-art methods, we show results for the original algorithms obtained from our own implementations, which are on par with the published results.

From the table, we see that our method consistently improves the performance over all baselines (\ie MT, MM and FM), especially in the low label regime on both CIFAR-10 \& -100. 
Particularly, our method achieves significant improvement over MT baseline, especially in the low label regime, up-to 12 points in case of 250 labels in CIFAR-10. 
Compared with the more competitive MM and FM baselines, our method obtains 6\% and 3\% improvement in case of 500 labels in CIFAR-100 (only 5 labels per class).
The results demonstrate that our method is complementary to the state-of-the-art methods in image classification and consistently boosts their performance in multiple settings.

\vspace{0.05cm}
\noindent \textbf{Classification on ImageNet.}
We also evaluate our method on ImageNet~\cite{deng2009imagenet} to verify that it performs well on larger and more complex image classification dataset. 
To make the training faster, we use the down-scaled ImageNet from \cite{Rebuffi2017LearningMV} that contains 1000 categories and 1.2 million images, each image is resized isotropically to have a shorter side of 72 pixels.
We randomly pick 5\% of the original training set as the labeled training set and the rest as the unlabelled set. 
We build our model on the state-of-the-art FixMatch~\cite{sohn2020fixmatch}, compare our method to FixMatch, MixMatch, standard supervised training on the labeled training set.
For all the methods, we use ResNet26~\cite{he2016deep} network as in~\cite{Rebuffi2017LearningMV}, train the networks from scratch, apply standard data augmentation such as random crop, flipping as in \cite{berthelot2019mixmatch} expect for FixMatch which incorporates RandAugment~\cite{cubuk2019randaugment} for strong augmentation.
We optimize all methods for 120 epochs by using SGD with a learning rate of 0.1 and momentum of 0.9, lower the learning rate by one tenth at every 30 epochs and report the median testing error of the last 20 epochs.

\vspace{0.05cm}
\noindent \textbf{Experiment 3.}
\Cref{tab:imagenet} depicts the test error for supervised learning (SL) using labels of full training set and SL using 5\% labels, the state-of-the-art semi-supervised learning methods, MixMatch and FixMatch and our method. 
While all the semi-supervised methods outperform the standard supervised training (SL) when using the same amount of labels, our method improves over FixMatch around 3\%.
The results demonstrate the efficacy of our method on large and complex data.

\begin{table}
	\centering
	\begin{tabular}{lcc}
	\toprule%
	method  & \#labels 	& Error (\%)\\%
	\midrule%
	SL &  100\% & 35.90 \\
	\midrule
	SL    & \multirow{4}{*}{5\%}  & 73.88\\%
	MM~\cite{berthelot2019mixmatch} &  	& 69.26\\%
	FM~\cite{sohn2020fixmatch} & & 59.55 \\
	{\bf FM-L2I (Ours)}      &          & {\bf 56.30}\\%
	\bottomrule%
	\end{tabular}%
	\caption{Top-1 error rate (\%) in image classification on ImageNet with 5\% supervised examples. FM-L2I denotes FixMatch with our method.}%
	\label{tab:imagenet}%
\end{table}%

\vspace{0.05cm}
\noindent \textbf{Text classification on IMDb.}
Apart from image data, we further evaluate our method on text data to show the generality of our approach. We evaluated our method on a sentiment analysis task using the IMDb dataset~\cite{maas2011learning}; this is a binary classification task.
The dataset contains 25,000 movie review texts for training and 25,000 for testing. 
Following the training and evaluation protocol of UDA~\cite{xie2019unsupervised}, we randomly sample 20 examples from training data and the rest of training examples and 50,000 unlabeled data are used as unlabeled data for evaluating all methods. 
We use the preprocessed data from UDA~\cite{xie2019unsupervised} and use the BERT model \cite{devlin2018bert}. The BERT model takes as input a sentence and performs text binary (positive or negative review) classification.
The model is initialized using the parameters from BERT~\cite{devlin2018bert} and is finetuned on IMDb as UDA.
Different from \cite{xie2019unsupervised}, we limit the input sequence length to 128 instead of 512 due to memory constraints in our hardware.
Thus we report results with a maximum sequence length of 128 for all the methods and a batch size of 8 and 24 for labeled and unlabeled respectively.



\begin{table}
	\centering
	\begin{tabular}{lc}
	\toprule%
	Method 	& Error (\%)\\%
	\midrule%
	SL     	                            &26.30 $\pm$ 1.09\\%
	UDA~\cite{xie2019unsupervised}   	& 9.82 $\pm$ 0.09\\%
	{\bf UDA-L2I (Ours)}                & {\bf 9.19 $\pm$ 0.10}\\%
	\bottomrule%
	\end{tabular}%
	\caption{Testing error rate (\%) in text classification on IMDb with only 20 supervised examples.}%
	\label{tab:imdb}%
\end{table}%

\vspace{0.05cm}
\noindent \textbf{Experiment 4.}
\Cref{tab:imdb} depicts the test error for the supervised learning baseline (SL), the state-of-the-art method (UDA \cite{xie2019unsupervised}) and our method. 
Our method obtains significantly better performance than the UDA (\ie 9.19 vs 9.82) which demonstrates that our method generalizes to semi-supervised problems in multiple domains.

\subsection{Ablation Study}
In this section, we describe ablation studies to further investigate the effect of 1) hold-out mini-batch size; 2) using joint or separate train and hold-out sets; 3) using an approximate update rule in \cref{eq:approx}; 4) modeling missing labels as model output or learnable parameters.

\vspace{0.05cm}
\noindent \textbf{Hold-out mini-batch size.}
In all the experiments, we use the same mini-batch size for the training and hold-out samples.
Here, our goal is to show that our method is robust to different hold-out mini-batch sizes.
To this end, we fix the training mini-batch size to 64 and evaluate our method (FM-L2) by varying the hold-out mini-batch size to 32, 64 and 128 in \cref{tab:batchsize}.
The results shows that 32 and 64 achieve good results in this setting, however, using 128 samples are slightly worse than 32 and 64.
\begin{table}[h]\label{tab:batchsize}
\centering
\resizebox{0.99\textwidth}{!}
{
	\begin{tabular}{lccc}
	\toprule
	Hold-out batch size & 32 & 64 & 128\\
	\midrule
	FM-L2I              & 44.67 $\pm$ 0.23 & 44.97 $\pm$ 0.13 & 45.78 $\pm$ 0.40\\
	\bottomrule
	\end{tabular}
}	
\vspace{-.3cm}
\caption{Test error (\%) in CIFAR-100 for various hold-out mini-batch sizes with 500 labels on CIFAR-100. FM-L2I denotes FixMatch with our method.}
\end{table}

\vspace{0.05cm}
\noindent \textbf{Joint training and hold-out sets.}
During training our method samples the hold-out mini-batch from the training set. 
Alternatively, we can split the training set to two mutually exclusive separate training and hold-out sets.
To this end, we randomly pick 60\% labels per class from the training set for training and 40\% as hold-out samples (\eg in case of 500 labels in CIFAR-100, we use 3 labels per class for training set and 2 labels per class for hold-out samples). 
We denote this as ``separate'' and evaluate our method trained in two settings in \Cref{tab:holdout}.
Note that we use the same total number of training and hold-out samples for both strategies.
The results show that using a joint set is a better strategy on average.

\begin{table}[h]\label{tab:holdout}
\centering
\resizebox{1.\textwidth}{!}
{
	\begin{tabular}{lccc}
	\toprule
	\# labels & 500 & 1000 & 2000\\
	\midrule
	FM-L2I (\textit{joint})    & {\bf 44.97 $\pm$ 0.13} & 39.21 $\pm$ 0.10 & {\bf 35.45 $\pm$ 0.10} \\
	FM-L2I (\textit{separate}) & 45.27 $\pm$ 0.03 & {\bf 39.20 $\pm$ 0.06} & 35.49 $\pm$ 0.15 \\
	\bottomrule
	\end{tabular}
}	
\vspace{-.3cm}
\caption{Test error rate (\%) in CIFAR-100 for joint and separate training and hold-out sets. Separate denotes mutually exclusive sets for training and hold-out samples and FM-L2I denotes FixMatch with our method.}
\end{table}

\vspace{0.05cm}
\noindent \textbf{Approximate update rule.}
Here we compare the effect of the exact and approximate update rules (see~\cref{eq:approx}) on AFLW, CIFAR-10 and CIFAR-100.
In particular, we use MT as our baseline in the experiments and denote the approximate update rule with ``approx'' in \cref{tab:approxalfw,tab:approx}.
We see that the approximate update rule performs better in AFLW and comparably in CIFAR datasets.
We reason that Hessian computation for high-dimensional tensors is  typically hard and inaccurate, thus the approximation presents a good tradeoff between accuracy and speed.
The approximation approximately leads to 50\% reduction in train time.
Note that we only show results for MT setting, as the exact update rule is computationally expensive for MM and FM.



\begin{table}[h]
	\centering
	\resizebox{1.\textwidth}{!}
	{
		\begin{tabular}{lccc}
			\toprule
			
			
			\#labels  & 91 (1\%) & 182 (2\%) & 455 (5\%) \\
			\midrule
			MT-L2I (\textit{exact})& 11.70 $\pm$ 0.03 & 10.49 $\pm$ 0.05 & 9.07 $\pm$ 0.03  \\
			MT-L2I (\textit{approx}) & {\bf 11.53 $\pm$ 0.02} & {\bf 10.32 $\pm$ 0.08} & {\bf 8.89 $\pm$ 0.08} \\
			\bottomrule
		\end{tabular}%
		}
		\vspace{-0.3cm}
		\caption{Testing mean square error on AFLW for the exact and approximate update rules.}
	\label{tab:approxalfw}%
\end{table}%

\begin{table}[h]
\centering
            \resizebox{1.\textwidth}{!}
  			{
            \begin{tabular}{lccc|ccc}

			& \multicolumn{3}{c}{CIFAR10} & \multicolumn{3}{c}{CIFAR100} \\
			\toprule
			
			\#labels        & 250   & 500    & 1000   & 500 & 1000 & 2000 \\
			\midrule	
			MT-L2I (\textit{exact}) & 27.72 &  {\bf 21.05}  &  14.93  & {\bf 69.57} & 59.25 & {\bf 51.68}  \\
			MT-L2I (\textit{approx}) & {\bf 25.97} & 21.13 & {\bf 15.76} & 70.58 & {\bf 58.96} & 51.95  \\
			\bottomrule
		\end{tabular}%
		}
			\vspace{-.3cm}
            \caption{Test error rate (\%) on CIFAR-10 and CIFAR-100 for the exact and approximate update rules.}
            \label{tab:approx}%
\end{table}

\vspace{0.05cm}
\noindent \textbf{Modeling missing labels.}
So far, we consider the missing labels as the output of $\psi$ and thus $\Psi_{\theta}$.
However, as discussed above, they can also be treated as learnable parameters.
Here we compare two strategies in landmark regression (see~\cref{tab:v1vsv2alfw}) and image classification tasks (see~\cref{tab:v1vsv2alfw}).
We denote the first and second setting with ``O'' and ``L'' respectively.
We see that both methods consistently achieve better performance than the corresponding baseline in the regression task. However, the second technique (L) is less effective on the CIFAR-10 dataset.
Modeling the missing labels as the model output is more effective on both the regression and classification tasks.



\begin{table}
	\centering
	\resizebox{.9\textwidth}{!}
	{
		\begin{tabular}{lccc}
			\toprule
			
			
			\#labels  & 91 (1\%) & 182 (2\%) & 455 (5\%) \\
			\midrule
			MT~\cite{tarvainen2017mean} & 12.80 $\pm$ 0.00 & 11.43 $\pm$ 0.02 & 9.46 $\pm$ 0.01 \\
			MT-L2I (O) & {\bf 11.53 $\pm$ 0.02} & \bf{10.32 $\pm$ 0.08} & \bf{8.89 $\pm$ 0.08} \\
			MT-L2I (L) & 11.78 $\pm$ 0.03 & 10.34 $\pm$ 0.09 & 9.02 $\pm$ 0.09 \\
			\bottomrule
		\end{tabular}%
		}
		\vspace{-0.3cm}
		\caption{Mean test error (\%) in AFLW when the missing labels are modeled as as model \textbf{O}utput and \textbf{L}earnable parameters.}
	\label{tab:v1vsv2alfw}%
\end{table}%


			

\begin{table}
\centering
            \resizebox{.9\textwidth}{!}
  			{
            \begin{tabular}{lccc}

			\multicolumn{4}{c}{CIFAR10}  \\
			\toprule
			
			\#labels           & 250   & 500    & 1000   \\
			\midrule
			FM~\cite{sohn2020fixmatch}      & 5.70 $\pm$ 0.04 & 5.78 $\pm$ 0.04 & 5.55 $\pm$ 0.07 \\
			FM-L2I (O) & {\bf 5.56 $\pm$ 0.03} & {\bf 5.70 $\pm$ 0.02} & {\bf 5.38 $\pm$ 0.02} \\
			FM-L2I (L) & 5.63 $\pm$ 0.03 & 5.74 $\pm$ 0.03 & 5.43 $\pm$ 0.08 \\
			\bottomrule\\
			\multicolumn{4}{c}{CIFAR100} \\
			\toprule
			\#labels & 500 & 1000 & 2000 \\
			\midrule
			FM~\cite{sohn2020fixmatch} & 48.32 $\pm$ 0.23 & 40.68 $\pm$ 0.10 & 35.52 $\pm$ 0.06 \\
			FM-L2I (O) & {\bf 44.97 $\pm$ 0.13} & {\bf 39.21 $\pm$ 0.10} & {\bf 35.45 $\pm$ 0.10} \\
			FM-L2I (L) & 48.00 $\pm$ 0.16 & 40.33 $\pm$ 0.03 & 35.53 $\pm$ 0.11 \\
			\bottomrule
		\end{tabular}%
		}
			\vspace{-.3cm}
            \caption{Test error rate (\%) in CIFAR-10 and CIFAR-100 using WideResNet-28-2 when the missing labels are modeled as model \textbf{O}utput and \textbf{L}earnable parameters.}
            \label{tab:v1vsv2cifar}%
\end{table}


\section{Conclusion}\label{s:conc}
In this work, we proposed a general semi-supervised learning framework that learns to impute labels of unlabeled data such that training a deep network on these labels improves its generalization ability.
Our method supports both classification and regression, and is easily integrated into several state-of-the-art semi-supervised methods.
We demonstrated that integrating our method results in significant performance gains over the baseline state-of-the-art methods, on challenging benchmarks for facial landmark detection, image and text classification, especially when the labeled data is scarce.
As future work, we plan to extend our method to semi-supervised learning for structured prediction tasks.



\bibliographystyle{plain}
\bibliography{refs}

\begin{IEEEbiography}[{\includegraphics[width=1in,height=1.25in,clip,keepaspectratio]{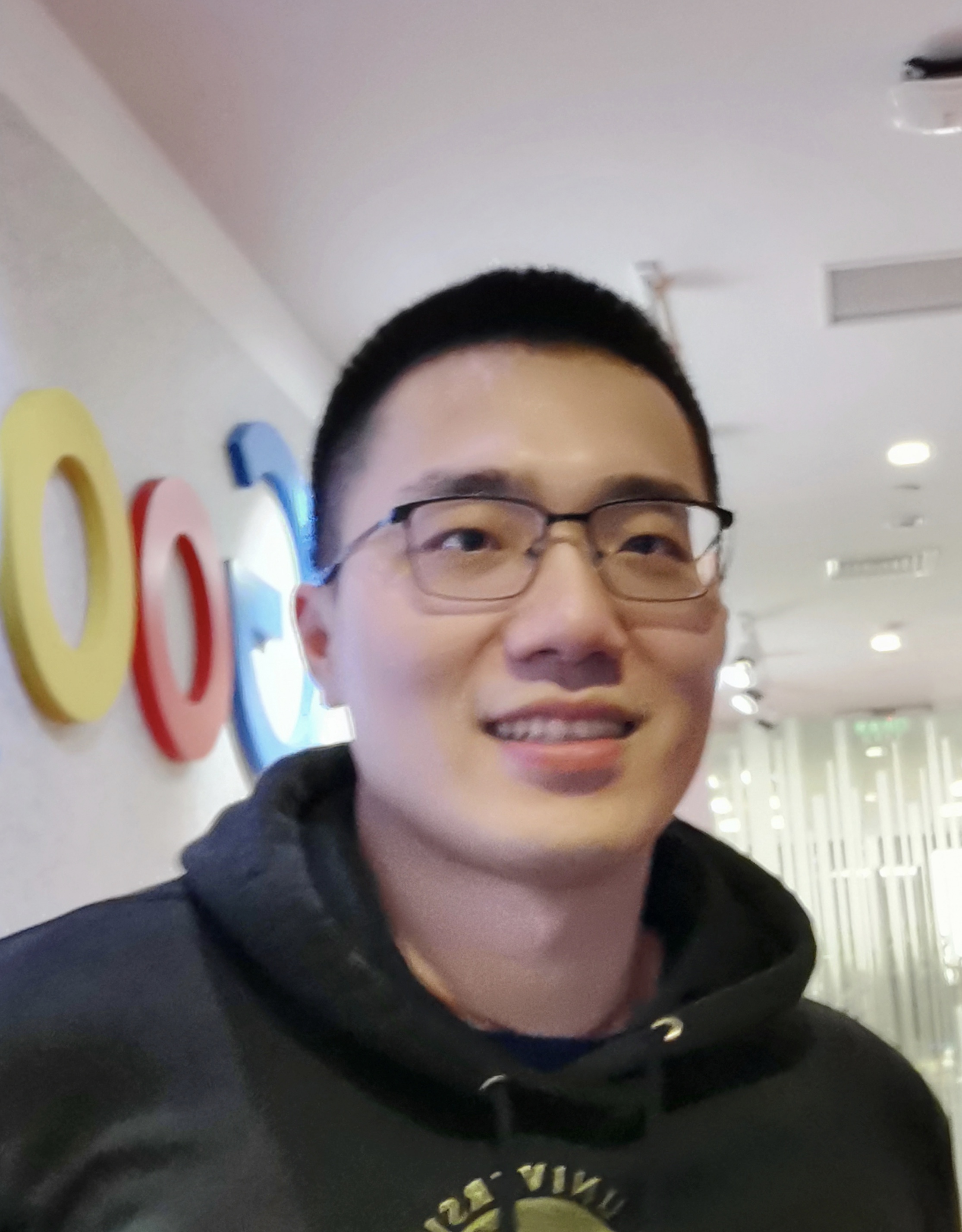}}]{Wei-Hong Li} is a PhD student in the School of Informatics at the University of Edinburgh. 
\end{IEEEbiography}

\begin{IEEEbiography}[{\includegraphics[width=1in,height=1.25in,clip,keepaspectratio]{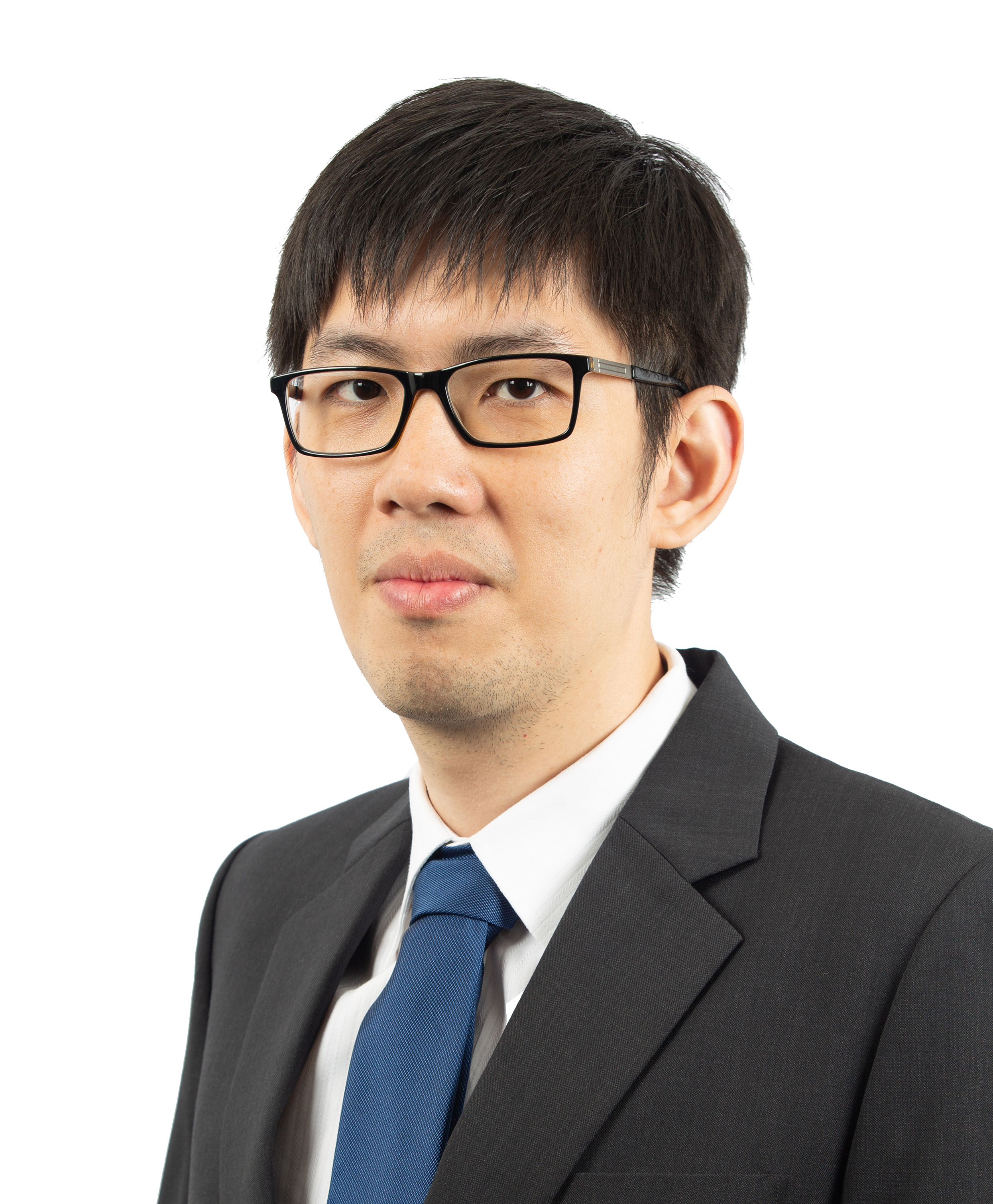}}]{Chuan-Sheng Foo} received his BS, MS and PhD from Stanford University. He currently leads a research group at the Institute for Infocomm Research, A*STAR, Singapore, which focuses on developing data-efficient deep learning algorithms that can learn from less labeled data. 
\end{IEEEbiography}


\begin{IEEEbiography}[{\includegraphics[width=1in,height=1.25in,clip,keepaspectratio]{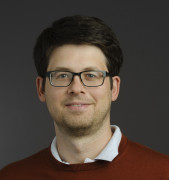}}]{Hakan Bilen} is an assistant professor in the School of Informatics at the University of Edinburgh. Previously he was a postdoc in Visual Geometry Group at the University of Oxford and received his PhD degree in Electrical Engineering at the KU Leuven in Belgium. 
\end{IEEEbiography}

\onecolumn
\newpage
\appendices
\section{Implementation details}\label{s:appendix}

In this section, we further explain the implementation details for the reported experiments that are conducted on AFLW and CIFAR-10/100.

\subsection{Regression experiments on AFLW}
\begin{figure}[h]
  \centering
  \includegraphics[width=.99\linewidth]{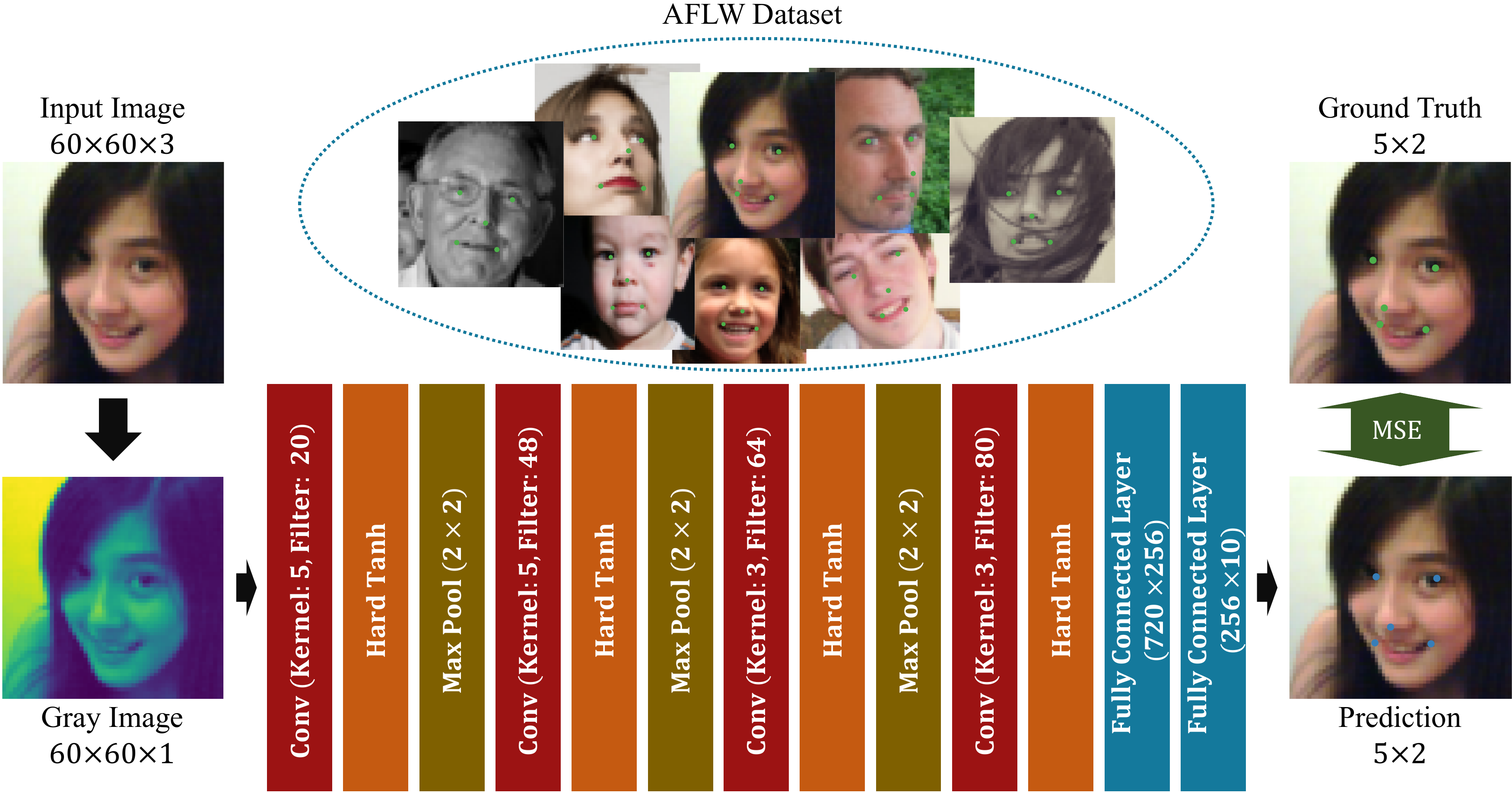}  
\caption{Illustration of the TCDCN.}\label{fig:TCDCN}
\end{figure}

\vspace{0.05cm}
\noindent \textbf{Experiment 1.}
We adopt the TCDCN proposed in \cite{zhang2015learning} as the network and SGD as optimizer. An illustration of the TCDCN's architecture is shown in \cref{fig:TCDCN}. 
We adapt the Pseudo Label (PL), Mean Teacher (MT) and the supervised learning methods as the baselines and our method is built on PL as well as MT. To estimate the loss on an unlabeled image, we firstly crop an image from the original image by moving the cropping window a random number of pixels. We then estimate the  location of landmarks on the augmented image and subtract the number of moving pixels, resulting in the pseudo label for the original image. We then apply MSE to the prediction of the original image and the pseudo labels to estimate the loss. We use the linear-schedule to update the unsupervised loss weight $\lambda$ to 3 in 9000 steps.

\subsection{Classification Experiments on CIFAR-10 \& -100}

For all experiments on both CIFAR-10 \& -100, 
we follow the optimizing strategy in \cite{berthelot2019mixmatch,sohn2020fixmatch}, i.e. we adopt Adam as the optimizer with fixed learning rate in CIFAR-10 and SGD as optimizer with initial learning rate of 0.03, momentum of 0.9, and a learning rate schedule of cosine learning rate decay in CIFAR-100 \cite{sohn2020fixmatch}. We evaluate the model using an exponential moving average of the learned models' parameters with a decay rate of 0.999. In addition, the weight decay is set to 0.0004 for all methods in CIFAR-10 and is 0.001 in CIFAR-100. We report the median error of the last 20 epochs for comparisons. For image preprocessing, we use standard data augmentation such as standard normalization and flipping, random crop as \cite{berthelot2019mixmatch} for all methods except FixMatch. In FixMatch (FM), we follow the original paper and use the RandAugment with the parameters reported in \cite{sohn2020fixmatch}. For our method (including modeling missing labels as \textbf{L}earnable parameters and model \textbf{O}utput), we impute the labels of unlabeled images by applying softmax on the augmentation of the original unlabeled image and use Mean Square Error (MSE) between prediction on the original image and pseudo labels for the unlabeled loss ($\ell^u$) as in \cite{berthelot2019mixmatch}.

\vspace{0.1cm}
\noindent \textbf{Experiment 2.}
We adopt the WideResNet-28-2 as the network for all methods as \cite{berthelot2019mixmatch}. The batch size on CIFAR-10 and CIFAR-100 is set to 64. We use the same consistency weight and ramp-up schedule (linear-schedule) to increase the unsupervised loss weight $\lambda$ as \cite{berthelot2019mixmatch}. The unsupervised loss weight is 1 for FixMatch as in \cite{sohn2020fixmatch}

\subsection{Classification on ImageNet}
We use a processed version of ImageNet~\cite{Rebuffi2017LearningMV} where each image has been resized isotropically to 72 $\times$ 72. In semi-supervised learning setting, we use 5\% training samples per class as labeled examples and the rest are as unlabeled samples. We use the ResNet26~\cite{he2016deep} network architecture as~\cite{Rebuffi2017LearningMV}. For SL, MM, we use standard augmentation techniques such as random crop, flipping as \cite{berthelot2019mixmatch}. For FixMatch, we use RandAugment~\cite{cubuk2019randaugment} with the same hyper-parameter in~\cite{sohn2020fixmatch} as strong augmentation. The batch size of labeled training mini batch is 256. We use the same batch size for unlabeled data for MM while the unlabeled batch size in FixMatch is 768. The unsupervised loss weight $\lambda$ is 100 for MixMatch and is 10 for FixMatch as in the original paper~\cite{sohn2020fixmatch}, respectively.

\vspace{0.05cm}
\noindent \textbf{Experiment 3.}
We optimize all method for 120 epochs by using SGD with a learning rate of 0.1 and nesterov. The momentum is 0.9 and weight decay is 0.0001. We scale the learning rate by 0.1 every 30 epochs and report the median testing error of the last 20 epochs.

\subsection{Text Classification on IMDb}
Following UDA~\cite{xie2019unsupervised}, the same 20 labeled examples are used for training and as hold-out samples, which means we \emph{do not} use additional labels for hold-out samples. Therefore, as UDA, we have 20 labeled training examples and 69,980 unlabeled samples for training. 

\vspace{0.05cm}
\noindent \textbf{Experiment 4.} 
We use the $\text{BERT}_{\text{BASE}}$~\cite{devlin2018bert} as the network and Adam as optimizer with a learning rate of $2\times10^{-5}$. The training batch size for labeled and unlabeled are 8 and 24 as UDA. The hold-out batch size is 8. The input sequence length is 128 in our work. We finetune the BERT model for 10k steps with the first 1k iterations being the warm-up phase.

\section{Analysis for one-layer network}\label{s:appendix2}
Here we analyze our method in case of an one-layer network parameterized by $\theta$ to perform logistic regression (\ie $\Phi_{\theta}(\bx)=\sigma(\theta^\top \bx)$ where $\sigma$ is sigmoid operation).
We focus on two cases, binary classification and regression for semi-supervised learning.
We show that the update rules in $\frac{\partial{C}^\mathcal{H}}{\partial \theta} = 
\frac{\partial C^{\mathcal{H}}}{\partial \Phi_{\theta^{\ast}}} \frac{\partial \Phi_{\theta^{\ast}}}{\partial \theta^{\ast}} \frac{\partial \theta^{\ast}}{\partial \psi }
\frac{\partial \psi}{\partial \theta }$ and $\frac{\partial{C}^\mathcal{H}}{\partial \bz} = 
\frac{\partial C^{\mathcal{H}}}{\partial \Phi_{\theta^{\ast}}} \frac{\partial \Phi_{\theta^{\ast}}}{\partial \theta^{\ast}} \frac{\partial \theta^{\ast}}{\partial \bz}$ are intuitive,
proportional to similarity between unlabeled and hold-out samples.

\subsection{Binary classification.}
During the training, given a set of training data $\mathcal{T}=\{\bx, y\}$, we consider using loss function $\ell(\cdot)=y \log(\sigma(\theta^\top \bx)) + (1-y) \log (1-\sigma(\theta^\top \bx))$. 
Regarding to the unlabeled samples $\mathcal{U} = \{\bx^u\}$, we define $z=\psi(\bx^u)=\sigma(\theta^\top(\bx^u+\eta))$ to impute label $z$ for $\bx^u$ and exploit the loss function $\ell^u(\cdot)=z \log(\sigma(\theta^\top \bx^u)) + (1-z) \log (1-\sigma(\theta^\top \bx^u))$ to regularize the consistency between the network's output $\sigma(\theta^\top\bx^u)$ and the imputed label $z=\sigma(\theta^\top(\bx^u+\eta))$.

We then detail how we solve OPT2 as below. At each iteration $t$,we first impute the label for unlabeled samples as $z=\psi(\bx^u)=\sigma(\theta^\top(\bx^u+\eta))$. We then compute the loss of unlabeled data and its derivative w.r.t $\theta$ to simulate a step of SGD as follow:
\begin{equation}
\begin{aligned}
\theta^{\ast}(\{z\}) &= \theta - \eta_{\theta} \nabla_{\theta} (C^{\mathcal{T}}+C^{\mathcal{U}}({z}))\\
&=\theta - \eta_{\theta} \nabla_{\theta} \left(\sum_{\bx, y\in\mathcal{T}}\ell(\Phi_\theta(\bx), y)+\sum_{\bx^u\in\mathcal{U}}\ell^u(\Phi_\theta(\bx^u), z)\right)\\
&= \theta - \eta_{\theta} \left(\sum_{\bx, y\in\mathcal{T}} \left((y-\sigma(\theta^{\top}\bx))\bx\right)+\sum_{\bx^u\in\mathcal{U}} \left((z-\sigma(\theta^{\top}\bx^u))\bx^u\right)\right).
\end{aligned}
\end{equation}

\vspace{0.05cm}
\noindent \textbf{Modeling missing labels as the model output.}
We wish to find $\{z\}$ via $\theta$ that minimizes $C^{\mathcal{H}}(\theta^{\ast}(\{z(\theta)\}))$. We apply the chain rule to $C^{\mathcal{H}}(\theta^{\ast}(\{z(\theta)\}))$ to obtain
\begin{equation}\label{eq:option2z}
\begin{aligned}
\frac{\partial C^{\mathcal{H}}}{\partial z}\frac{\partial z}{\partial \theta} &= \left(\frac{\partial C^{\mathcal{H}}}{\partial \Phi_{\theta^{\ast}}} \frac{\partial \Phi_{\theta^{\ast}}}{\partial \theta^{\ast}}\right)^\top \frac{\partial \theta^{\ast}}{\partial z}\frac{\partial z}{\partial \psi}\frac{\partial \psi}{\partial \theta}\\
&=\sum_{\bx, y\in\mathcal{H}}\left(\frac{\partial \ell(\bx, y)}{\partial \sigma({\theta^{\ast}}^\top\bx)} \frac{\partial \sigma({\theta^{\ast}}^\top\bx)}{\partial \theta^{\ast}}\right)^\top \frac{\partial \left(\theta - \eta_{\theta} \left(\sum_{\bx, y\in\mathcal{T}} \left((y-\sigma(\theta^{\top}\bx))\bx\right)+\sum_{\bx^u\in\mathcal{U}} \left((z-\sigma(\theta^{\top}\bx^u))\bx^u\right)\right)\right)}{\partial z} \\
&\sigma(\theta^\top(\bx^u+\eta))(1-\sigma(\theta^\top(\bx^u+\eta)))(\bx^u+\eta)\\
&=\sum_{\bx, y\in\mathcal{H}}(y - \sigma({\theta^{\ast}}^\top\bx))\bx^\top \left(\frac{\partial \theta}{\partial z}-\eta_{\theta}\bx^u\right)\sigma(\theta^\top(\bx^u+\eta))(1-\sigma(\theta^\top(\bx^u+\eta)))(\bx^u+\eta).
\end{aligned}
\end{equation}
Since $\theta$ is not the function of $z$, we can rewrite \cref{eq:option2z} as:
\begin{equation}\label{eq:option2zfinal}
\begin{aligned}
\frac{\partial C^{\mathcal{H}}}{\partial z}\frac{\partial z}{\partial \theta} = \eta_{\theta}\sum_{\bx\in\mathcal{H}}(\sigma({\theta^{\ast}}^\top\bx)-y)\bx^\top\bx^u \sigma(\theta^\top(\bx^u+\eta))(1-\sigma(\theta^\top(\bx^u+\eta)))(\bx^u+\eta).
\end{aligned}
\end{equation}

\vspace{0.05cm}
\noindent \textbf{Modeling missing labels as learnable parameters.}
So far, we treat $\{z\}$ as the output of $\psi$, which is parameterized by $\theta$. Alternatively, $z$ can also be the learnable parameters and initialize them with the imputing function $\psi$. The imputing process can also be achieved by optimizing $z$ by one SGD step:
\begin{equation}
z^{\ast} = z - \eta_z \nabla_z C^{\mathcal{H}}(\theta^{\ast}(\{z\})),
\end{equation}
where
\begin{equation}\label{eq:option1z}
\begin{aligned}
\frac{\partial C^{\mathcal{H}}}{\partial z} &= \left(\frac{\partial C^{\mathcal{H}}}{\partial \Phi_{\theta^{\ast}}} \frac{\partial \Phi_{\theta^{\ast}}}{\partial \theta^{\ast}}\right)^\top \frac{\partial \theta^{\ast}}{\partial z}\\
&=\sum_{\bx, y\in\mathcal{H}}\left(\frac{\partial \ell(\bx, y)}{\partial \sigma({\theta^{\ast}}^\top\bx)} \frac{\partial \sigma({\theta^{\ast}}^\top\bx)}{\partial \theta^{\ast}}\right)^\top \frac{\partial \left(\theta - \eta_{\theta} \left(\sum_{\bx, y\in\mathcal{T}} \left((y-\sigma(\theta^{\top}\bx))\bx\right)+\sum_{\bx^u\in\mathcal{U}} \left((z-\sigma(\theta^{\top}\bx^u))\bx^u\right)\right)\right)}{\partial z} \\
&=\sum_{\bx, y\in\mathcal{H}}(y - \sigma({\theta^{\ast}}^\top\bx))\bx^\top \left(\frac{\partial \theta}{\partial z}-\eta_{\theta}\bx^u\right)\\
&=\eta_{\theta}\sum_{\bx,y\in\mathcal{H}}(\sigma({\theta^{\ast}}^\top\bx)-y)\bx^\top\bx^u.
\end{aligned}
\end{equation}


From \cref{eq:option2zfinal} and \cref{eq:option1z},
the update rule is proportional to the similarity (\ie $\bx^\top\bx^u$) between the unlabeled and hold-out samples, and to the loss ($(\sigma({\theta^{\ast}}^\top\bx)-y)$) between the output of the network and the ground-truth labels. Intuitively, the algorithm updates the imputed label towards the majority class of the incorrectly predicted samples on the hold-out set, weighted by the similarity of the unlabeled sample to the respective hold-out sample; correctly predicted hold-out samples do not contribute to the update.

\subsection{Regression.} Similar to the binary classification mentioned above, we use the same setting with the mean square error loss function (\ie $\ell(\cdot)=||\theta^\top\bx-y||^2$) for regression problem. To optimize OPT2, we first compute the updated model by one SGD step:
\begin{equation}
\begin{aligned}
\theta^{\ast}(\{z\})&=\theta - \eta_{\theta} \nabla_{\theta}(C^{\mathcal{T}}+C^{\mathcal{U}}(z))\\
&=\theta - \eta_{\theta} \nabla_{\theta} \left(\sum_{\bx,y\in\mathcal{T}}\ell(\Phi_\theta(\bx), y)+\sum_{\bx^u\in\mathcal{U}}\ell^u(\Phi_\theta(\bx^u), z)\right) \\
&=\theta - \eta_{\theta} \left(\sum_{\bx,y\in\mathcal{T}} \frac{\partial ||\theta^\top\bx - y||^2}{\partial \theta}+\sum_{\bx^u\in\mathcal{U}} \frac{\partial ||\theta^\top\bx^u - z||^2}{\partial \theta}\right)  \\
&=\theta - 2\eta_{\theta} \left(\sum_{\bx,y\in\mathcal{T}} (\theta^{\top}\bx-y)\bx+\sum_{\bx^u\in\mathcal{U}} (\theta^{\top}\bx^{u}-z)\bx^u\right).
\end{aligned}
\end{equation}

\vspace{0.05cm}
\noindent \textbf{Modeling missing labels as the model output.}
We wish to find $\{z\}$ via $\theta$ that minimizes $C^{\mathcal{H}}(\theta^{\ast}(\{z(\theta)\}))$. We apply the chain rule to $C^{\mathcal{H}}(\theta^{\ast}(\{z(\theta)\}))$ to obtain
\begin{equation}\label{eq:option2zmse}
\begin{aligned}
\frac{\partial C^{\mathcal{H}}}{\partial z}\frac{\partial z}{\partial \theta} &= \left(\frac{\partial C^{\mathcal{H}}}{\partial \Phi_{\theta^{\ast}}} \frac{\partial \Phi_{\theta^{\ast}}}{\partial \theta^{\ast}}\right)^\top \frac{\partial \theta^{\ast}}{\partial z}\frac{\partial z}{\partial \psi}\frac{\partial \psi}{\partial \theta}\\
&=\sum_{\bx, y\in\mathcal{H}}\left(\frac{\partial \ell(\bx, y)}{\partial {\theta^{\ast}}^\top\bx} \frac{\partial {\theta^{\ast}}^\top\bx}{\partial \theta^{\ast}}\right)^\top\frac{\partial \theta^{\ast}}{\partial z} (\bx^u+\eta)\\
&=\sum_{\bx, y\in\mathcal{H}}4\eta_{\theta}({\theta^{\ast}}^{\top}\bx-y)\bx^\top\bx^u(\bx^u+\eta).
\end{aligned}
\end{equation}

\vspace{0.05cm}
\noindent \textbf{Modeling missing labels as learnable parameters.}
So far, we treat $\{z\}$ as the output of $\psi$, which is parameterized by $\theta$. Alternatively, $z$ can also be the learnable parameters and initialize them with the imputing function $\psi$. The imputing process can also be achieved by optimizing $z$ by one SGD step:
\begin{equation}
z^{\ast} = z - \eta_z \nabla_z C^{\mathcal{H}}(\theta^{\ast}(\{z\})),
\end{equation}
where
\begin{equation}\label{eq:derivativemse}
\begin{aligned}
\frac{\partial C^{\mathcal{H}}}{\partial z} &= \left(\frac{\partial C^{\mathcal{H}}}{\partial \Phi_{\theta^{\ast}}} \frac{\partial \Phi_{\theta^{\ast}}}{\partial \theta^{\ast}}\right)^\top \frac{\partial \theta^{\ast}}{\partial z}\\
&=\sum_{\bx, y\in\mathcal{H}}\left(\frac{\partial \ell(\bx, y)}{\partial {\theta^{\ast}}^\top\bx} \frac{\partial {\theta^{\ast}}^\top\bx}{\partial \theta^{\ast}}\right)^\top\frac{\partial \theta^{\ast}}{\partial z}\\
&=\sum_{\bx, y\in\mathcal{H}}4\eta_{\theta}({\theta^{\ast}}^{\top}\bx-y)\bx^\top\bx^u. \\
\end{aligned}
\end{equation}

\end{document}